\pdfoutput=1
\documentclass[runningheads]{llncs}
\usepackage{graphicx}

\usepackage{tikz}
\usepackage{comment}
\usepackage{amsmath,amssymb} 
\usepackage{color}

\usepackage{booktabs}
\usepackage{makecell}
\usepackage[labelfont=bf]{caption}
\usepackage{wrapfig}
\usepackage{bbm}
\usepackage{multirow}
\usepackage{floatrow}
\usepackage{xcolor,colortbl}

\usepackage{microtype}

\newfloatcommand{capbtabbox}{table}[][\FBwidth]

\begin{document}
\pagestyle{headings}
\mainmatter
\def\ECCVSubNumber{1114}

\title{
CramNet: Camera-Radar Fusion \\ with Ray-Constrained Cross-Attention \\ for Robust 3D Object Detection}

\titlerunning{CramNet}
%
\author{Jyh-Jing Hwang, Henrik Kretzschmar, Joshua Manela, Sean Rafferty, \\[1ex] Nicholas Armstrong-Crews, Tiffany Chen, Dragomir Anguelov}
\authorrunning{Hwang et al.}
%
\institute{Waymo}

\maketitle

\begin{abstract}

Robust 3D object detection is critical for safe autonomous driving.
Camera and radar sensors are synergistic as they capture complementary information and work well under different environmental conditions.
Fusing camera and radar data is challenging, however, as each of the sensors lacks information along a perpendicular axis, that is, depth is unknown to camera and elevation is unknown to radar.
We propose the camera-radar matching network CramNet, an efficient approach to fuse the sensor readings from camera and radar in a joint 3D~space.
To leverage radar range measurements for better camera depth predictions, we propose a novel ray-constrained cross-attention mechanism that resolves the ambiguity in the geometric correspondences between camera features and radar features.
Our method supports training with sensor modality dropout, which leads to robust 3D~object detection, even when a camera or radar sensor suddenly malfunctions on a vehicle.
We demonstrate the effectiveness of our fusion approach through extensive experiments on the RADIATE dataset, one of the few large-scale datasets that provide radar radio frequency imagery.
A camera-only variant of our method achieves competitive performance in monocular 3D~object detection on the Waymo Open Dataset.

\end{abstract}

\begin{figure}[t]
\centering
  \includegraphics[width=1.0\linewidth]{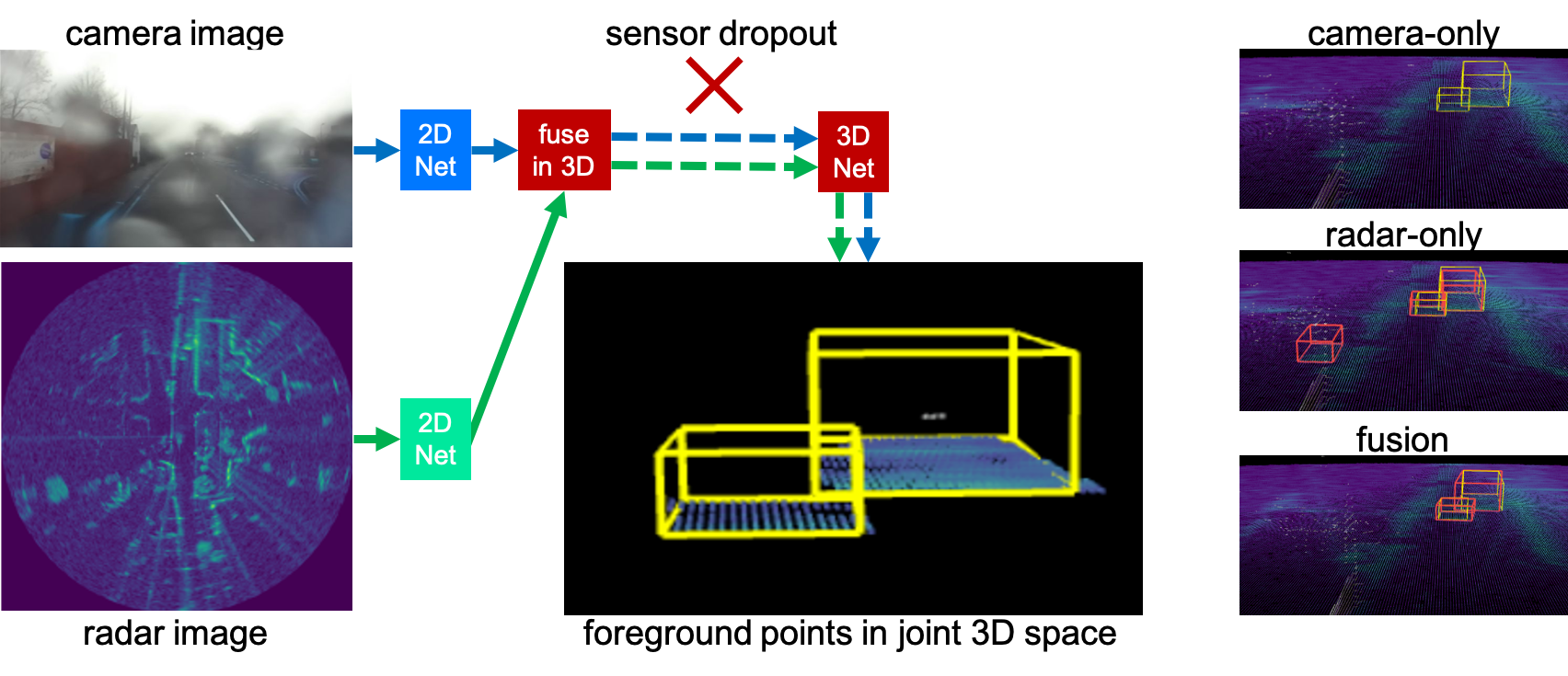}
  \vspace{-20 pt}
  \caption{Our approach takes as input a camera image (top left) and a radar RF image (bottom left). The model then predicts foreground segmentation for both native 2D~representations before projecting the foreground points with features into a joint 3D~space (middle bottom) for sensor fusion. Finally, the method runs sparse convolutions in the joint space for 3D~object detection. The network architecture naturally supports training with sensor dropout. This allows the resulting model to cope with sensor failures at inference time as it can run on camera only and radar only input depending on which sensors are available. \vspace{-20 pt}
  } 
  \label{fig:teaser}
\end{figure}

\section{Introduction}
\label{sec:intro}

3D~object detection that is robust to different weather conditions and sensor failures is critical for safe autonomous driving.
Fusion between camera and radar sensors stands out as they are both relatively resistant to various weather conditions~\cite{bijelic2020seeing} compared to the popular lidar sensor~\cite{bijelic2018benchmark}.
A fusion design that naturally accepts single-sensor failures (lidar, radar, or camera or radar) is thus desired and boosts safety in an autonomous driving system (Figure~\ref{fig:teaser}).

Most sensor fusion research has focused on fusion between lidar and another sensor~\cite{qi2018frustum,wang2019frustum,chen2017multi,huang2020epnet,vora2020pointpainting,wang2021pointaugmenting,liang2018deep,huang2020epnet,piergiovanni20214d,yang2020radarnet,shah2020liranet}  because lidar provides complete geometric information, i.e., azimuth, range, and elevation.
Sparse correspondences between lidar and another sensor is thus well defined, making lidar an ideal carrier for fusion.
On the other hand, even though camera and radar sensors are lighter and cheaper, consume less power, and endure longer than lidar, camera-radar fusion is understudied.
Camera-radar fusion is especially challenging as each sensor lacks information along one perpendicular axis: depth unknown for camera and elevation unknown for emerging imaging radar, as summarized in Table~\ref{tab:sensor}.
Radar produces radio frequency (RF) imagery that encodes the environment approximately in the bird's-eye view (BEV) with various noise patterns, an example shown in Figure~\ref{fig:teaser}.
As a result, camera data (in perspective view) and radar data (in BEV) form many-to-many mappings and the exact matching is unclear from geometry alone.


To solve the matching problem, we consider three possible schemes for fusion:
\textbf{(1) Perspective view primary}~\cite{qi2018frustum}:  This scheme implies we trust the depth reasoning from the perspective view.  One can project camera pixels to their 3D locations with depth estimates and find their vertical nearest neighbors of corresponding radar points.  If depth is unknown, one can project a pixel along a ray in 3D and perform matching.
\textbf{(2) Bird’s-eye view primary}~\cite{vora2020pointpainting}:  This scheme implies we trust the elevation reasoning from the bird’s-eye view.  However, since it’s difficult to predict elevation from radar imagery directly, one might borrow elevation information from the map.  Hence, the inferred elevation for radar is sometimes inaccurate, resulting in rare usage unless LiDAR is available. 
\textbf{(3) Cross-view matching}~\cite{kim2020grif}: This scheme implies we perform matching in a joint 3D space.  For example, one can use supplementary information (map or camera depth estimation) to upgrade camera and radar 2D image pixels to 3D point clouds (with some uncertainty) and perform matching between point clouds directly.  This is supposedly the most powerful scheme if we can properly handle uncertainties.  Our architecture is designed to enable this matching scheme, hence we name it CramNet (Camera and RAdar Matching Network).

\begin{table}[t]
\centering
  \resizebox{0.9\linewidth}{!}{%
  \begin{tabular}{l | c |c |c |c | c}
    \toprule
    Sensor & Azimuth & Range & Elevation & Resistance to weather & 3D detection literature \\
    \midrule
    Camera & \checkmark & x & \checkmark & medium & abundant \\
    Radar & \checkmark & \checkmark & x$^*$ & high & scarce \\
    Lidar & \checkmark & \checkmark & \checkmark & low & abundant \\
    \bottomrule
  \end{tabular}}
  \caption{Characteristics of major sensors commonly used for autonomous driving.  Both camera and radar tend to be less affected by inclement weather compared to lidar scanners.  However, whereas regular camera does not directly measure range, radar does not measure elevation. This poses a unique challenge for fusing camera and radar readings as the geometric correspondences between the two sensors are underconstrained. Overall, camera-radar fusion is still underexplored in the literature. $^*$Although there exists radars with elevation, this paper focuses on planar radar which, at the moment, is more common for automotive radar.  %
  \vspace{-18 pt}}
  \label{tab:sensor}
\end{table}

Since the effectiveness of projecting into 3D space heavily relies on accurate camera depth estimates, we propose a ray-constrained cross-attention mechanism to leverage radar for better depth estimation.
The idea is to match radar responses along each camera ray emitted from a pixel.
The correct projection should be the locations where radar senses reflections.
Our architecture is further designed to accept sensor failures naturally.
As shown in Figure~\ref{fig:teaser}, the model is able to operate even when one of the modalities is corrupted during inference.
To this end, we incorporate sensor dropout~\cite{chen2017multi,wang2020makes} in the point cloud fusion stage during training to boost the sensor robustness. 

We summarize the contributions of this paper as follows:
\vspace{-6pt}
\begin{enumerate}
\item We present a camera-radar fusion architecture for 3D~object detection that is flexible enough to fall back to a single sensor modality in the event of a sensor failure.
\item We demonstrate that the sensor fusion model effectively leverages data from both sensors as the model outperforms both the camera-only and the radar-only variants significantly.
\item We propose a ray-constrained cross-attention mechanism that leverages the range measurements from radar to improve camera depth estimates, leading to improved detection performance.
\item We incorporate sensor dropout during training to further improve the accuracy and the robustness of camera-radar 3D~object detection.
\item We demonstrate state-of-the-art radar-only and camera-radar detection performance on the RADIATE dataset~\cite{sheeny2021radiate} and competitive camera-only detection performance on the Waymo Open Dataset~\cite{Sun2020CVPR}.
\end{enumerate}

\section{Related Work}
\label{sec:work}

\noindent \textbf{Camera-based 3D object detection.}
Monocular camera 3D object detection is first approached by directly extending 2D detection architectures and incorporating geometric relationships between the 2D perspective view and 3D space~\cite{chen2016monocular,mousavian20173d,brazil2019m3d,simonelli2019disentangling,liu2020reinforced,simonelli2020towards,chen2020monopair,li2020rtm3d}.  Utilizing pixel-wise depth maps as an additional input shows improved results, either for lifting detected boxes~\cite{manhardt2019roi,shi2020distance} or projecting image pixels into 3D point clouds~\cite{wang2019pseudo,ma2020rethinking,you2019pseudo,ding2020learning,weng2019monocular} (also known as Pseudo-LiDAR~\cite{wang2019pseudo}).  More recently, another camp of methods emerge to be promising, i.e., projecting intermediate features into BEV grid features along the projection ray without explicitly forming 3D point clouds~\cite{roddick2018orthographic,srivastava2019learning,reading2021categorical,li2022bevformer}.

The BEV grid methods benefit from naturally expressing the 3D~projection uncertainty along the depth dimension.  However, these methods suffer from significantly increased compute requirements as the detection range expands. In contrast, we model the depth uncertainty through sampling along the projection ray and consulting radar features for more accurate range signals.  This also enables the adoption of foreground extraction that allows a balanced trade-off between detection range and computation.

\noindent \textbf{Radar-based 3D object detection.}
Frequency modulated continuous wave (FMCW) radar is usually presented by two kinds of data representations, i.e., radio frequency (RF) images and radar points. 
The RF images are generated from the raw radar signals using a series of fast Fourier transforms that encode a wide variety of sensing context whereas the radar points are derived from these RF images through a peak detection algorithm, such as Constant False Alarm Rate (CFAR) algorithm~\cite{richards2010principles}.
The downside of the radar points is that recall is imperfect and the contextual information of radar returns is lost, with only the range, azimuth and doppler information retained.
As a result, radar points are not suitable for effective single modality object detection~\cite{schumann2018semantic,qi2017pointnet++}, which is why most works use this data format only to foster fusion~\cite{bijelic2020seeing,kim2020grif,nobis2021radar,nabati2021centerfusion}.
On the other hand, the RF images maintain rich environmental context information and even complete object motion information to enable a deep learning model to understand the semantic meaning of a scene~\cite{major2019vehicle,sheeny2021radiate}.
Our work is therefore built upon radar RF images and can produce reasonable 3D object detection predictions with radar-only inputs. 


\noindent \textbf{Sensor fusion for 3D object detection.}
Sensor fusion for 3D object detection has been studied extensively using lidar and camera.  The reasons are twofold: 1) Lidar scans provide comprehensive representations in 3D for inferring correspondences between sensors, and 2) camera images contain more semantic information to further boost the recognition ability.  Various directions have been explored, such as image detection in 2D before projecting into frustums~\cite{qi2018frustum,wang2019frustum}, two-stage frameworks with object-centric modality fusion~\cite{chen2017multi,huang2020epnet,li2022deepfusion}, image feature-based lidar point decoration~\cite{vora2020pointpainting,wang2021pointaugmenting}, or multi-level fusion~\cite{liang2018deep,huang2020epnet,piergiovanni20214d}.  Since sparse correspondences between camera and lidar are well defined, fusion is mostly focused on integrating information rather than matching points from different sensors.

As a result, these fusion techniques are not directly applicable to camera-radar fusion where associations are underconstrained.  Early work, Lim et al.~\cite{lim2019radar}, applies feature fusion directly between camera and radar features without any geometric considerations. Recently, more works tend to leverage camera models and geometry for association.  For example, CenterFusion~\cite{nabati2021centerfusion} creates camera object proposal frustums to associate radar features and GRIF Net~\cite{kim2020grif} projects 3D RoI to camera perspective and radar BEV to associate features. 
Our model, on the other hand, fuses camera-radar data in a joint 3D space with the flexibility to perform 3D detection with either single modality, leading to increased robustness.

\section{CramNet for Robust 3D Object Detection}
\label{sec:method}

We describe the overall architecture for camera-radar fusion in Section~\ref{sec:method_arch}.
%
%
In Section~\ref{sec:method_transformer}, We then introduce a ray-constrained cross-attention mechanism to leverage radar for better camera 3D point localization.
Finally, we propose sensor dropout that can be integrated seamlessly into the architecture in Section~\ref{sec:method_dropout} to further improve the robustness of 3D object detection.

\begin{figure}[t]
    \centering
    \includegraphics[width=\linewidth]{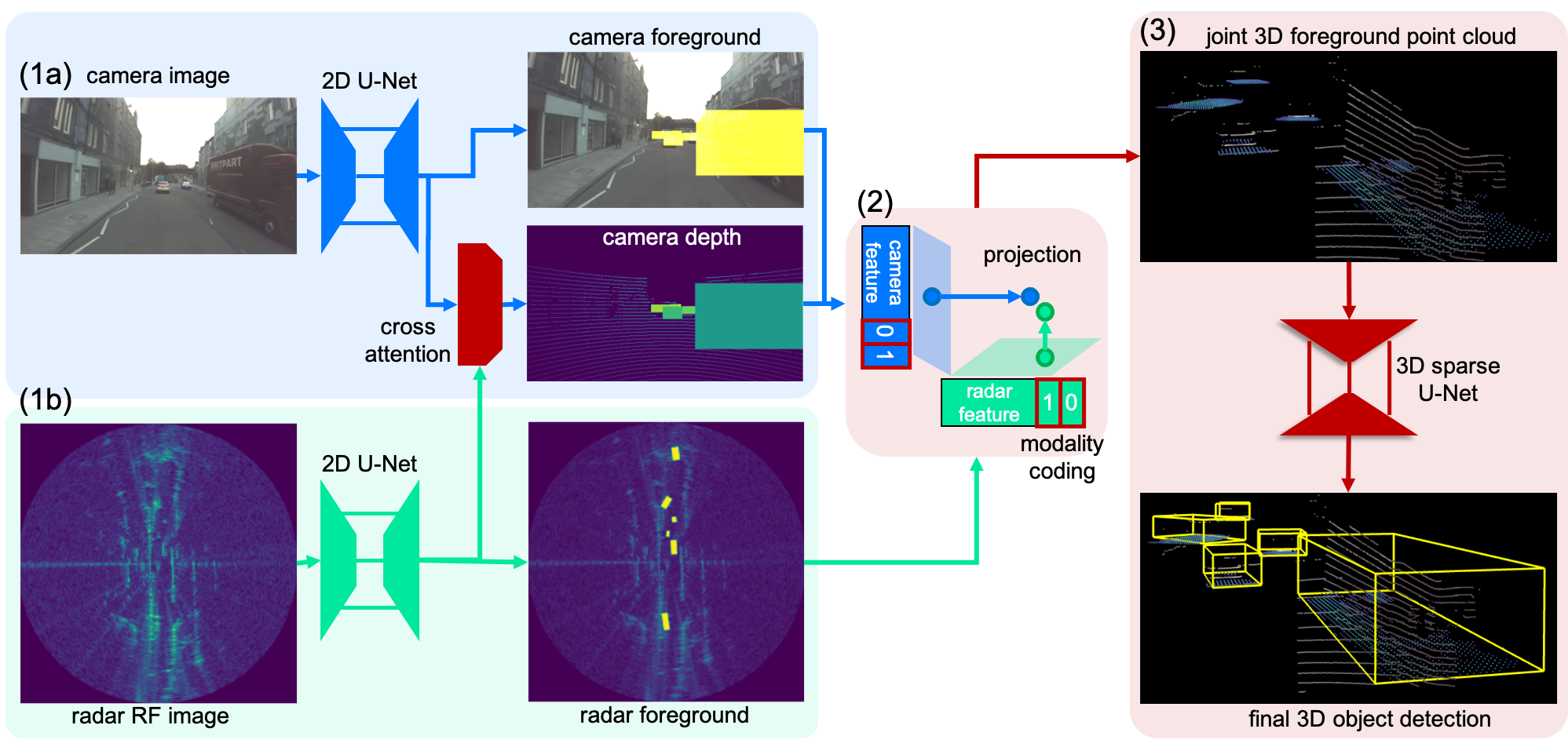}
    \caption{Architecture overview.  Our method can be partitioned into three stages: (1a) camera 2D foreground segmentation and depth estimation, (1b) radar 2D foreground segmentation, (2) projection from 2D to 3D and subsequent point cloud fusion, and (3) 3D foreground point cloud object detection. 
    The cross-attention mechanism modifies the camera depth estimation by consulting radar features, as further illustrated in Figure~\ref{fig:transformer}.
    The modality coding module appends a camera or radar binary code to the features that are fed into the 3D~stage, enabling sensor dropout and enhancing robustness.
    We depict the camera stream in blue, the radar stream in green, and the fused stream in red.
    \vspace{-12 pt}}
    \label{fig:architecture}
\end{figure}

\subsection{Overall Architecture}
\label{sec:method_arch}

Our model architecture, in Figure~\ref{fig:architecture}, is inspired by Range Sparse Net (RSN)~\cite{sun2021rsn}, which is an efficient two-stage lidar-based object detection framework.
The RSN framework takes input of perspective range images, segments perspective foreground pixels, extracts 3D (BEV) features on foreground regions using sparse convolution~\cite{yan2018second}, and performs CenterNet-style~\cite{zhou2019objects} detection.
We adapt the framework for camera-radar fusion and the overall architecture can be partitioned into three stages: (1) 2D foreground segmentation, (2) 2D to 3D  projection and point cloud fusion, and (3) 3D foreground point cloud detection. 

\textbf{Stage 1: 2D foreground segmentation.} 
The goal of this stage is to perform efficient foreground segmentation for native dense representations from two modalities. This allows us to restrict the expensive 3D operations to foreground points.
The network takes as input a pair of camera images $\pmb{I}_C$ and radar RF images $\pmb{I}_R$.
We then employ two identical lightweight U-Nets~\cite{ronneberger2015u} to extract 2D  features and predict foreground segmentation masks for each modality, $\pmb{F}_C$ and $\pmb{F}_R$, respectively.
For camera image feature extraction, one can also adopt a more powerful, multi-scale feature extractor, such as a feature pyramid network~\cite{lin2017feature}.
The detailed design of the U-Net can be found in the supplementary.

To train such a segmentation network (for both camera and radar), we use the 2D projection of 3D bounding box labels as ground truth -- a pixel belongs to the foreground class if it falls inside any of the projected 2D boxes.
This might introduce some noise as background pixels sometimes fall within a box, but we find that this noise is insignificant in practice.
We then apply a pixel-wise focal loss~\cite{lin2017focal} to classify each pixel:
\begin{equation}
    L_\text{seg} = \frac{-1}{N} \Big( \sum_{i\in \pmb{F}}  (1-p_{i})^{\gamma_s} \log (p_{i}) + \sum_{i\in \pmb{B}} p_{i}^{\gamma_s} \log (1-p_{i}) \Big),
\end{equation}
where $N$ is the total number of pixels, $\pmb{F}$ and $\pmb{B}$ are the sets of foreground and background pixels, and $p_i$ is the model's estimated probability of foreground for pixel $i$.
The hyperparameter $\gamma_s$ controls the penalty reduction.
A pixel with foreground score higher than $\tau$ will be selected.
Since the 3D stage can resolve false positives, whereas false negatives cannot be recovered, we typically set a low value for $\tau$ to attain high recall.

\textbf{Stage 2: 2D to 3D projection and point cloud fusion.} 
Once we obtain the foreground pixels, we project them into 3D for the following 3D stage.
For the camera projection, we predict a depth value for each pixel from the same U-Net with additional convolutional layers.
The depth ground truth is obtained by projecting lidar points to the camera view and overlaying them with depth values from projected ground truth 3D boxes. 
The use of depth from ground truth boxes is to enable 3D detection where lidar data alone is insufficient.
This is especially true outside of lidar range, as well as when lidar points are deteriorated due to weather.
We train the depth estimation using pixelwise L2 losses on valid regions, or $L_\text{depth}$.
The camera projection relies on the camera model, i.e., the intrinsics and extrinsics, with depth to infer the 3D location of each pixel.

For radar projection, we use the radar model to transform radar BEV points to 3D using the sensor height as elevation.
If map is available, the road elevation can be used to offset this value to handle non-planar scenes like hills.

There are several options to combine the camera and radar 3D point clouds.
One plausible choice is to select one modality as a major sensor and gather features from the other modality.
This is usually how researchers fuse lidar with other sensors~\cite{vora2020pointpainting}.
However, the drawback is obvious: the major sensor is a single point of failure.
Instead, we directly place two point clouds in a joint 3D space.
We align the feature dimensions of both modalities and append a modality code to the feature so that the 3D network can leverage the multi-modality information easily.
The major benefit is to enable robust detection especially when one modality fails to perform.

\textbf{Stage 3: 3D foreground point cloud detection.} 
We apply dynamic voxelization~\cite{zhou2020end} on the fused foreground point cloud, whose features are then encoded into sparse voxel features.
A 2D or 3D sparse convolution network~\cite{graham2017submanifold} (for pillar style~\cite{lang2019pointpillars},
or 3D voxelization, respectively) is applied on the sparse voxels.
The network details can be found in the supplementary. 

We follow RSN~\cite{sun2021rsn} for CenterNet-style~\cite{zhou2019objects} 3D box regression.
We calculate a ground truth objectness heatmap for every point $x\in \mathbb{R}^3$:
$h(x) = \max \{ \exp(-\frac{|| x - c|| - || x_c - c|| }{ \sigma^2}) \mid c \in C(x) \}$ where $C(x)$ is the set of centers of boxes containing $x$, $x_c$ is the closest point to box center $c$, and $\sigma$ is a constant.
In other words, the objectness of a point is inversely related to its distance to the closest box center.
We train the network to predict a heatmap using a focal loss~\cite{lin2017focal}:
\begin{align}
\nonumber L_\text{hm} =& \frac{-1}{N} \sum_{x} \Big( 
\big(1-\tilde{h}(x)\big)^{\gamma_h}  \log \big(\tilde{h}(x)\big) \mathbbm{1}(h>1-\epsilon_h) + \\
& \big(1-h(x)\big)^\alpha_h 
\tilde{h}(x)^{\gamma_h}  \log \big(1-\tilde{h}(x)\big) \mathbbm{1}(h\le 1-\epsilon_h)
\Big),
\end{align}
where $\mathbbm{1}(\cdot)$ is the indicator function, $h$ and $\tilde{h}$ are the ground truth and predicted heat map, $(1-\epsilon_h)$ decides the threshold for ground truth objectness, and $\alpha_h$ and $\gamma_h$ are hyperparameters in the focal loss.

The 3D boxes are parameterized as $\pmb{b} = (b_x, b_y, b_z, l, w, h, \theta)$ where $b_x, b_y, b_z$ are the offsets of a 3D box center relative to a voxel center, and $l, w, h, \theta$ are the length, width, height, and heading of a box.
All the box parameters are trained with smooth L1 losses except for the heading that is trained with a bin loss~\cite{shi2019pointrcnn}.
An additional IoU loss~\cite{zhou2019iou} is employed for better accuracy.
The box regression loss is as follows:
\begin{equation}
    L_\text{box} = \frac{1}{B} \sum_i \Big(
    L_\text{SmoothL1}(\pmb{b}_i \backslash \theta_i - \tilde{\pmb{b}}_i \backslash \tilde{\theta}_i) + 
    L_\text{bin}(\theta_i, \tilde{\theta}_i) + L_{\text{IoU}i}
    \Big),
\end{equation}
where $B$ is the total number of boxes with ground truth heatmap value greater than a threshold $\tau_\text{hm}$, and $\pmb{b}_i$ and $\tilde{\pmb{b}}_i$ denote the ground truth and prediction for box $\pmb{b}_i$, respectively; the same for $\theta_i$.
For more details on heatmap and box regression, we refer interested readers to RSN~\cite{sun2021rsn}.

We train the fusion network end-to-end with losses summarized as:
\begin{equation}
    L = \lambda_\text{seg} L_\text{seg} + \lambda_\text{depth} L_\text{depth} + \lambda_\text{hm} L_\text{hm} + L_\text{box},
\end{equation}
where $\lambda_*$ are hyperparameters for the respective loss weighting.

\subsection{Ray-Constrained Cross-Attention}
\label{sec:method_transformer}

\begin{figure}[t]
    \centering
    \includegraphics[width=\linewidth]{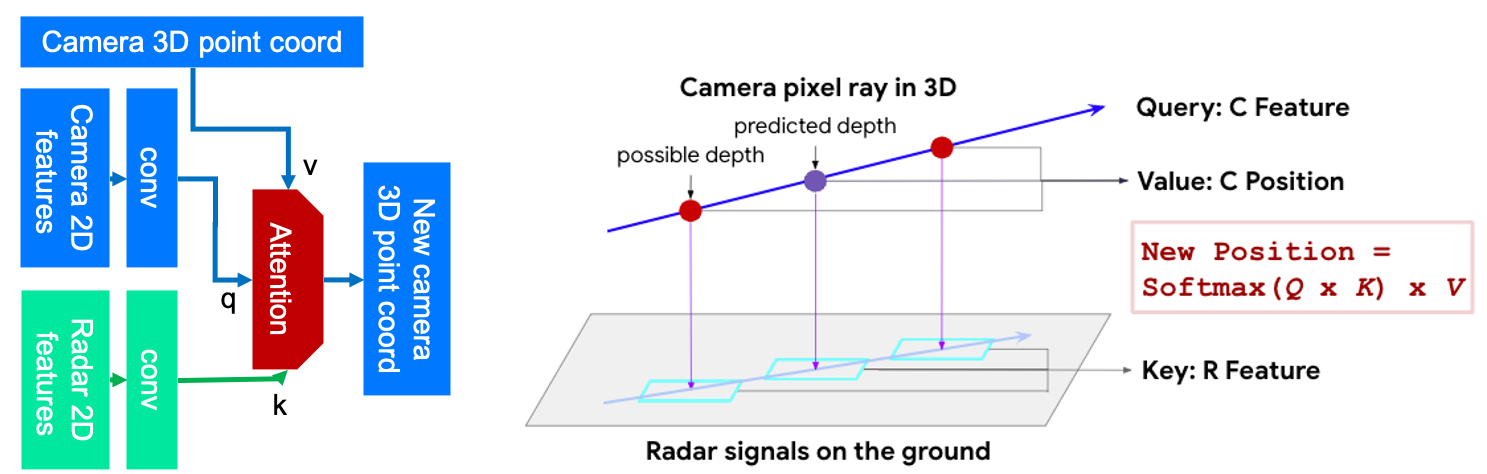}
    \caption{The proposed ray-constrained cross-attention mechanism resolves the  ambiguity in the geometric correspondences between camera features and radar features. Following the Transformer~\cite{vaswani2017attention}, we take camera features as queries and radar features as keys to transform 3D camera points as values. %
    \vspace{-12 pt}}
    \label{fig:transformer}
\end{figure}

It is widely known~\cite{park2021pseudo,wang2019pseudo} that camera-based 3D object detection relies heavily on accurate depth estimation, either explicitly or implicitly through BEV representations.
Luckily, for camera-radar fusion, we don't have to rely solely on camera to infer depth as radar provides relatively accurate range estimates.
To utilize the complementary sensing directions, we propose a ray-constrained cross attention mechanism to leverage radar for improving camera 3D point localization, illustrated in Figure~\ref{fig:transformer}.

Our observation is that an optimal 3D location for each foreground camera  pixel usually accompanies a corresponding peak response from radar.
Thus we propose to consult radar features along a camera 3D ray, emitted from each pixel, to rectify the camera 3D point location after projection.
Since there is infinite possible locations, we perform sampling along the ray, centered at the initial depth estimation.
The final 3D location is decided by matching between camera and radar features among these sampled locations.

We denote the projected camera 3D location from a depth estimate $\tilde{d}_i$ for pixel $i$ as $M(\tilde{d}_i)$. 
We sample $s$ points farther and closer around the estimated location respectively, or $M(\tilde{d}_i \pm \epsilon \times k)$, where $\epsilon$ is a hyperparameter for depth error, and $k$ ranges from 1 to $s$.
We denote this set of 3D locations as $\tilde{\pmb{M}}_i \in \mathbb{R}^{(2s+1) \times 3}$.
We gather closest radar features for every sampled location, denoted as $\pmb{\psi}_{Ri} \in \mathbb{R}^{(2s+1) \times d}$.
Likewise, we denote the camera feature for a pixel $i$ as $\pmb{\psi}_{Ci} \in \mathbb{R}^{1 \times d}$.
The final camera 3D point location $M_i\in \mathbb{R}^{1\times3}$ for pixel $i$ can thus be obtained using a cross-attention formulation~\cite{vaswani2017attention}:
\begin{equation}
    M_i = \text{softmax}(\frac{\pmb{\psi}_{Ci} \pmb{\psi}_{Ri}^T}{\sqrt{d}}) \tilde{\pmb{M}}_i.
\end{equation}
To relate to the naming convention in attention~\cite{vaswani2017attention}, we use the camera feature $\pmb{\psi}_{Ci}$ as a query, a set of radar features $\pmb{\psi}_{Ri}$ as keys, and the sampled 3D locations $\tilde{\pmb{M}}_i$ as values.
Therefore, the final location is calculated by matching the query with the most active keys, associated with the respective values.

Notably, this design is computationally efficient.
The time complexity is asymptotically proportional to $(N \times d \times s)$, where $N$ is the number of (foreground) pixels.
Since $s$ is a small constant, this operation is as cheap as a conv layer.

\subsection{Sensor Dropout}
\label{sec:method_dropout}

One appealing property of this architecture is the independence of each sensor.
We can perform camera-only or radar-only 3D object detection with the same architecture when one modality is unavailable.
This is desired in practice as one never knows when a sensor might be unavailable due to various situations, e.g., occlusions, weather, or sensor failure.

To enhance the model ability to handle sensor failures, we incorporate a sensor dropout mechanism~\cite{chen2017multi,wang2020makes} during training.
With a probability $P_\text{drop}$, we randomly drop out the entire set of point features of camera $\pmb{\psi}_C$  or radar $\pmb{\psi}_R$, or
%
%
\begin{align}
\nonumber \pmb{X}_C &= \mathbbm{1}(r_1 \ge P_\text{drop}) \ \pmb{X}_C  + \mathbbm{1}(r_1 < P_\text{drop} \land r_2 \ge 0.5) \ \pmb{0}  \\
\pmb{X}_R &= \mathbbm{1}(r_1 \ge P_\text{drop}) \ \pmb{X}_R  + \mathbbm{1}(r_1 < P_\text{drop} \land r_2 < 0.5) \ \pmb{0}
\end{align}
where $\pmb{0}$ is a zero matrix and $r_1$ and $r_2$ are uniform random numbers in [0, 1].  Note that camera and radar features won't be dropped out at the same time.  We use $p_\text{drop}=0.2$ in our experiments.

The reason why we choose to mask out 3D point features instead of input data directly is that we can still train the cross-attention with proper 2D features normally.
If radar sensor is corrupted during inference and produces noisy 2D features, it results in a uniform attention map inside cross-attention and little effect on 3D camera point locations.

\section{Experiments}
\label{sec:exp}

We present experiments on the RADIATE~\cite{sheeny2021radiate} and Waymo Open~\cite{Sun2020CVPR} datasets to verify the efficacy of our proposed CramNet model.
We introduce the settings in Section~\ref{sec:exp_dataset} and \ref{sec:exp_impl}.
We include the main results, ablation studies, robustness tests, and visualization on the RADIATE dataset in Section~\ref{sec:exp_result}, \ref{sec:exp_ablation}, \ref{sec:exp_robust}, and \ref{sec:exp_visual}, respectively.
We also present our camera-only results on the Waymo Open dataset in Section~\ref{sec:exp_open}.
More ablation studies can be found in the supplementary.

\subsection{Dataset and Evaluation}
\label{sec:exp_dataset}

\noindent \textbf{RADIATE dataset.}
We evaluate our method on the challenging RADIATE dataset~\cite{sheeny2021radiate}. This dataset features radar sensor data collected for scene understanding for safe autonomous driving in various weather conditions, including sunny, night, rainy, foggy, and snowy.
The dataset includes 3~hours of annotated radar imagery with more than 200k~labeled objects for 8~categories.
These properties make the RADIATE dataset one of the few public datasets that contain high-resolution radar data along with a large number of ground truth labels for road actors.
While the dataset provides high-quality radar data, the quality of its camera and LiDAR data is not comparable to that of other autonomous driving datasets, such as the Waymo Open Dataset~\cite{Sun2020CVPR}.
This shortcoming, however, makes the evaluation of the robustness of our proposed sensor fusion algorithm even more compelling.
In all of our experiments, we train the models on the training set that contains both good and bad weather conditions, and we evaluate the resulting models on the standard validation set.

\noindent \textbf{Evaluation.}
The (pseudo) 3D~labels in the RADIATE dataset are 2D~BEV labels with assumed heights for each category. We therefore report our 3D detection results in terms of BEV~AP to align with the baselines, unless otherwise noted.
We follow the proposed evaluation in the dataset and define the category ``vehicle'' to encompass the six categories ``car'', ``van'', ``truck'', ``bus'', ``motorbike'', and ``bicycle''.
The final BEV/3D~AP numbers are therefore weighted sums of the objects from these categories.
For all radar and fusion experiments, we evaluate the performance on the region that is captured by both the cameras and the radar sensors, up to the radar range of 100~meters.
For all camera-only experiments, we exclude labels that do not contain any LiDAR points.
The motivation for this is that camera depth estimates beyond the LiDAR supervision (up to 70~meters) tend to be inaccurate.

\begin{table}[t]
  \centering
  \resizebox{0.99\linewidth}{!}{%
  \begin{tabular}{l || c |c c c c c c c c}
    \toprule
    Method & \ Overall \  &
    \makecell{Sunny \\ (Parked)}  &
    \makecell{Overcast \\ (Motorway)} &
    \makecell{Sun/OC \\ (Urban)} &
    \makecell{Night \\ (Motorway)} & 
    \makecell{Rain \\ (Suburban)} & 
    \makecell{Fog \\ (Suburban)} &
    \makecell{Snow \\ (Surburban)} \\
    \midrule
    Baseline~\cite{sheeny2021radiate} & 46.55 & 79.72 & 44.23 & 35.45 & 64.29 & 31.96 & 51.22 & 8.14 \\
    \hline
    CramNet-C* & 23.66 & 67.98 & 6.50 & 23.43 & 2.24 & 17.69 & 9.50  & 0.12 \\ 
    CramNet-R & 56.19 & 83.58 & 37.65 & 48.33 & 60.38 & 42.86 & 71.11 & {\bf 15.84} \\
    CramNet & {\bf 62.07} & {\bf 96.68} & {\bf 50.49} & {\bf 52.25} & {\bf 79.56} & {\bf 57.90} & {\bf 85.26} & 8.89  \\ 
    \bottomrule
  \end{tabular}}
  \caption{Main results evaluated in BEV AP (\%) on the RADIATE dataset~\cite{sheeny2021radiate}. CramNet-C (*notes evaluation on camera/lidar-specific labels), CramNet-R, and CramNet denote our camera-only, radar-only, and fusion models, respectively.  Our final model outperforms the baseline Faster R-CNN~\cite{sheeny2021radiate} by $16$ percentage points, the camera-only variant by $38$ points, and the radar-only variant by $6$ points. These large gains validate the efficacy of our proposed sensor fusion model.
  \vspace{-30 pt}}
  \label{tab:radiate_main}
\end{table}

\subsection{Implementation Details}
\label{sec:exp_impl}


\noindent \textbf{Hyperparameters.}
CramNet follows the implementation of RSN~\cite{sun2021rsn}.
The sparse convolution implementation is also similar to \cite{yan2018second}.
The input camera and radar RF images are both normalized to be in [0, 1].
The foreground score cutoff is set to 0.15, the segmentation loss weight is set to 400, and the depth loss weight is set to 20.
For cross-attention, we sample 1 point closer and farther around the predicted depth location, with 10\% error. 
The voxelization region is [-100m, 100m] $\times$ [-100m, 100m] $\times$ [-5m, 5m] with 0.2 meter voxel sizes.
In the heatmap computation, $\sigma_h$ is set to 1.0, the heatmap loss weight is set to 4 and threshold $\epsilon_h$ are set to 0.2.
We use 12~bins in the heading bin loss for heading regression.

\noindent \textbf{Training and inference.}
We train CramNet from scratch end-to-end using the Adam optimizer~\cite{kingma2014adam} on Tesla V100 GPUs.
The models are trained with 5 batches on 8 GPUs.
We use a cosine learning rate decay, where we set the initial learning rate to 0.006, with 1k~warm-up steps starting at 0.003 and 50k~steps in total.
%
%
We use layer normalization~\cite{ba2016layer} instead of batch normalization~\cite{ioffe2015batch} in the 3D network for the number of foreground points varies among different scenes.
We do not perform 2D data augmentation but adopt two 3D data augmentation strategies, namely, random flipping along the x-axis and a global rotation around the z-axis, with a random angle from [-$\pi$/4, $\pi$/4] on the selected foreground points.

\subsection{Performance on the RADIATE dataset}
\label{sec:exp_result}


We evaluate the performance of our method on the RADIATE dataset~\cite{sheeny2021radiate} and summarize the results in Table~\ref{tab:radiate_main}.
We report the BEV AP at a 0.5~IoU threshold to align with the baseline proposed in the RADIATE dataset~\cite{sheeny2021radiate}.
The baseline runs a Faster R-CNN detector with a ResNet-101 backbone on radar RF images.

Our radar-only variant, CramNet-R, outperforms the baseline by a large margin, $\sim 10$ percentage points in AP.
Our two-stage framework effectively filters out radar noise in the segmentation stage to focus inference on the remaining radar signals in subsequent stages.
Our camera-only variant, CramNet-C, performs the worst. Several factors may contribute to the poor performance. First, adverse weather affects the cameras more than the radar sensors, which is exacerbated by the lack of wipers mounted on the vehicles. Second, the effective range of the LiDAR sensors, which we use for camera depth supervision at training time, tends to drop from 70 meters in clear weather to about 40 meters in adverse weather, whereas we have labeled ground truth boxes within a range of 100 meters.
Overall, we observe that the short sensing range and the sparsity of the points prevent the model from learning accurate camera depth estimation, resulting in poor camera-only 3D detection performance.

Our proposed fusion model, CramNet, equipped with ray-constrained cross attention and sensor dropout, outperforms the baseline BEV AP by $16$ percentage points, the camera-only variant by $38$ points, and the radar-only variant by $6$ points.
These large gains validate the efficacy of our proposed sensor fusion model.
In the next sections, we study the performance of our method in more detail.


\subsection{Ablation Study}
\label{sec:exp_ablation}

\begin{table}[!ht]
\resizebox{0.325\linewidth}{!}{%
  \begin{tabular}{c c | c }
    \toprule
    Attention \ & \ Dropout \ & \ BEV AP \\
    \midrule
     &  & 56.19  \\
    \checkmark &  & 60.20 \\
     & \checkmark & 61.23 \\
    \checkmark & \checkmark & 62.07 \\
    \bottomrule
  \end{tabular}}
\hspace{20pt}
\centering
\resizebox{0.5\linewidth}{!}{%
  \begin{tabular}{c | c | c | c }
    \toprule
    \makecell{Radar Intensity \\ Threshold} & \# of Points & BEV AP & Degradation \\
    \midrule
    None & - & 60.20 & - \\
    0.25 & 70K & 50.90 & -15.45\% \\
    0.5 & 2K & 17.81 & -70.42\% \\
    \bottomrule
  \end{tabular}}
\caption{Ablation study on CramNet on the RADIATE dataset~\cite{sheeny2021radiate}. Left: The cross-attention and sensor dropout both improve over the vanilla fusion model by $4$ to $5$ points in AP.  Putting them together yields the final fusion model with the best performance. Right: We simulate the radar sparse signals by setting the intensity thresholds to 0.25 or 0.5, resulting in $\sim$70K or 2K points, respectively. As a result, our model performance is degraded relatively by 15\% to 70\%. This confirms radar RF imagery contains critical information for 3D detection.
}
\label{tab:radiate_ablation}
\end{table}

\noindent \textbf{Effects of ray-constrained cross-attention and dropout.}
We experiment the fusion model with different settings to enable/disable ray-constrained cross-attention and sensor dropout mechanisms.
The experimental results are summarized in Table~\ref{tab:radiate_ablation} (left).
The cross-attention and sensor dropout both improve over the vanilla fusion model by $4$ to $5$ points in BEV AP.
Putting them together yields our final fusion model, achieving the performance of $62.07\%$ AP. 

\noindent \textbf{Effects of sampling radar points.}
Most of the camera-radar fusion methods, such as GRIF Net~\cite{kim2020grif} and CenterFusion~\cite{nabati2021centerfusion}, perform experiments on the nuScenes dataset~\cite{caesar2020nuscenes} that contains only sparse radar points, at the scale of hundreds of points in a scene.
The resulted radar-only model usually performs poorly, such as 25.5\% AP reported in~\cite{kim2020grif}.
On the other hand, our model is specifically designed to perform either single modality effectively by taking as input the RF images instead of sparse points.

To quantitatively study how the sparsity of radar signals affects the performance, we filter RF images with varying intensity thresholds, as summarized in Table~\ref{tab:radiate_ablation} (right).
We set the intensity thresholds to 0.25 or 0.5, resulting in $\sim$70K or 2K points, respectively, which are already denser than sparse radar points available on  nuScenes~\cite{caesar2020nuscenes} or SeeingThroughFog~\cite{bijelic2020seeing} datasets.
As a result, our model performance is degraded relatively by 15\% to 70\%.
We conclude that radar RF imagery contains critical information for effective 3D object detection.

%
%
%
%
%
%

%
%

\subsection{Detection Robustness}
\label{sec:exp_robust}

\begin{figure}[t!]
\begin{floatrow}
\begin{tabular}{l || c }
    \toprule
    Dropout Location \ & \ BEV AP  \\
    \midrule
    Normal & 57.00 \\
    Input & 55.78 \\
    Point Cloud & 58.64 \\
    Point Feature & 61.23 \\
    \bottomrule
\end{tabular}
\ffigbox{%
  \includegraphics[width=1.1\linewidth]{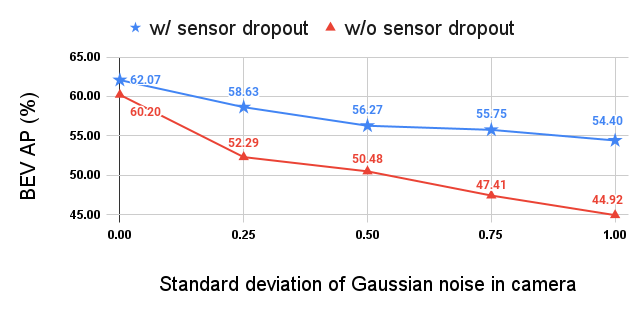}
  \vspace{-50pt}
}{}
\end{floatrow}
\caption{\textbf{Left}: Analysis on different sensor dropout strategies. Masking out point features yields the best performance.  Two possible benefits for this dropout location:  1) Reduce the 3D network reliance on features, which are disrupted the most given sensor noise. 2) Remain smooth training of 2D feature extractors and cross-attention.  \textbf{Right}: Analysis on model performance degradation on corrupted data.  We add varying degrees of Gaussian white noises to corrupt camera images and evaluate the performance.  Our fusion model trained with sensor dropout greatly outperforms the one without by 2 to 10 percentage points in BEV AP.  This demonstrates that sensor dropout can drastically enhance sensor robustness.
}
\label{fig:dropout}
\end{figure}

  %

Detection robustness against sensor deterioration is critical for safe autonomous driving.
In this section, we study the effects of our proposed sensor dropout with ablation study and corrupted sensor data.

\noindent \textbf{Where to drop out sensor data?}
Dropout is a popular technique for training neural network models~\cite{srivastava2014dropout}.
It is usually applied on a layer to randomly mask out neuron activations.
We experiment on various places to drop out sensor data and summarize them in Table~\ref{fig:dropout} (left).
The `normal' dropout applies the conventional dropout on point cloud features, regardless of which sensor the points are from.
This conventional dropout does not provide benefits in either overall performance or in bad weather conditions.
The `input' dropout randomly masks out a sensor (radar or camera) entirely.
The `point cloud' dropout randomly masks out the 3D points from one sensor entirely.
The `point feature' dropout randomly masks out the initial point cloud features from one sensor entirely but leaves the point cloud positions intact.
As the numbers dictate, masking out point features yields the best performance.
Two possible benefits for this dropout location:
(1) Reduce the 3D network reliance on features, which are disrupted the most due to sensor noise.
(2) Remain training of 2D feature extractors and cross-attention.
As such, we conclude dropping out sensor point features randomly is the most effective.

\noindent \textbf{Sensor dropout improves robustness against input corruption.}
We study how the corruption of sensors will affect the performance with and without sensor dropout and summarize the experiments in Table~\ref{fig:dropout} (right).
For this purpose, we add random Gaussian white noise with varying standard deviation to corrupt the camera images to different degrees.
We evaluate the fusion model on the corrupted data, with or without sensor dropout during training.
The experimental results show that our fusion model trained with sensor dropout greatly outperforms the one without by 2 to 10 percentage points in BEV AP.
This study demonstrates that sensor dropout can drastically enhance sensor robustness.

\subsection{Camera-only CramNet on Waymo Open Dataset}
\label{sec:exp_open}

Our radar-only and camera-radar fusion models perform strongly on the RADIATE dataset~\cite{sheeny2021radiate}.
However, the camera-only model suffers from the poor image quality and adverse weather conditions in the dataset.
Since we do not have access to another public dataset that contains radar RF imagery, we evaluate our camera-only model performance on the Waymo Open Dataset~\cite{Sun2020CVPR}, as summarized in Table~\ref{tab:exp_open}.
%
%
We report 3D AP/APH with 0.7 IoU threshold on the LEVEL\_1 difficulty in Table~\ref{tab:exp_open}.
Our camera-only model, CramNet-C, achieves competitive performance.
More details can be found in the supplementary.

\begin{table}[t]
\centering
\resizebox{1.0\linewidth}{!}{%
\begin{tabular}{l|c c c c|c c c c}
  \toprule
  Method & \ 3D AP \  & \ 0 - 30m \ & \ 30 - 50m \ & \ 50m - $\infty$\  & \ 3D APH\ & \ 0 - 30m\ & \ 30 - 50m\ & \ 50m - $\infty$\ \\
  \midrule
  M3D-RPN~\cite{brazil2019m3d} & 0.35 & 1.12 & 0.18 & 0.02 & 0.34 & 1.10 & 0.18 & 0.02 \\
  CaDDN~\cite{reading2021categorical} & \textbf{5.03} & 14.54 & \textbf{1.47} & 0.10 & \textbf{4.99} & 14.43 & \textbf{1.45} & 0.10 \\
  CramNet-C & 4.14 & \textbf{15.46} & 1.20 & \textbf{0.15} &
  4.10 & \textbf{15.31} & 1.19 & \textbf{0.13}\\
  \bottomrule
\end{tabular}}
\caption{Camera-only 3D detection results on the Waymo Open Dataset~\cite{Sun2020CVPR} validation set on the vehicle class, evaluated in terms of 3D~AP/APH at 0.7~IoU on the LEVEL\_1 difficulty.  Baseline numbers are from \cite{reading2021categorical}. Our camera-only model, CramNet-C, achieves competitive performance among state-of-the-art. 
%
}
\label{tab:exp_open}
\end{table}

\subsection{Visualization}
\label{sec:exp_visual}

\begin{figure}[t]
    \centering
    \includegraphics[width=\linewidth]{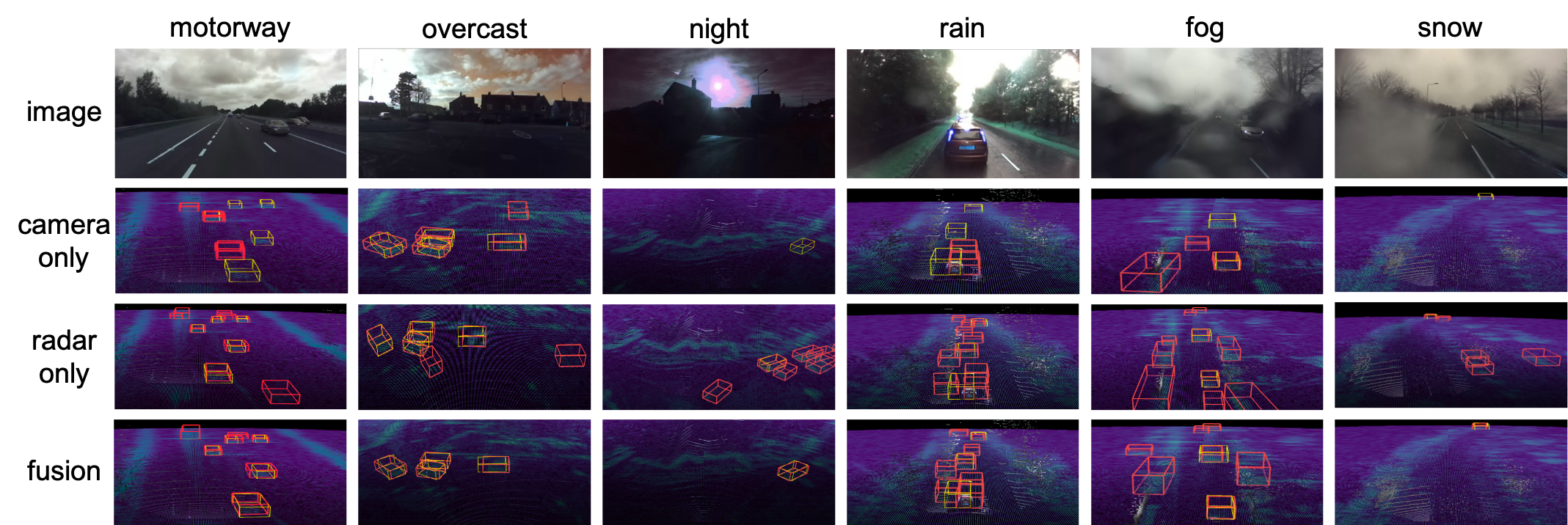}
    \caption{Visual comparison between CramNet-C, CramNet-R, and CramNet from 6 scenarios.  We visualize the predicted boxes in red and the ground truth boxes in yellow with projected radar and camera pixels.   Whereas the camera-only model tends to miss detections and predict inaccurate localization, the radar-only model suffers from false positives.  Our camera-radar fusion model combines the advantages of the two and produces the most accurate predictions.
    }
    \label{fig:exp_visual}
\end{figure}

We present the visual comparisons between our camera-only, radar-only, and fusion models in Figure~\ref{fig:exp_visual}.  
Since the camera visibility is severely reduced due to either underexposure or adverse weather, the camera-only model tends to miss detection and the predicted localization tends to be inaccurate. 
In contrast, the radar-only model suffers from false positives due to lack of appearance features from RF images. 
Overall, our camera-radar fusion model combines the advantages from the two and produces the most accurate predictions.

\section{Conclusion}
\label{sec:disc}

We introduced a camera-radar sensor fusion approach for robust 3D~object detection for autonomous driving.
The method relies on a ray-constrained cross-attention mechanism to leverage the range measurements from radar to improve camera depth estimates.
%
Training with sensor dropout allows the method to fall back to a single modality when one of the sensors malfunctions.
We present experiments on the RADIATE dataset and the Waymo Open Dataset.

\textbf{Limitations.} 
Whereas a camera pixel corresponds to a ray, a (range, azimuth) radar reading corresponds to an arc in 3D space.
Intersecting a camera ray and a radar arc yields their correspondence.
We approximate the radar arc as a pillar, that is, we assume that the radar points are at the same elevation as the sensor.
This assumption works well in practice when most objects are at a similar elevation as the sensor.
We currently use the RF images in Cartesian coordinates, which may be suboptimal as the radar natively operates in polar coordinates.
We will explore a polar convolutional network design and radar-specific spherical voxelization in future work.

\clearpage
%
%
\bibliographystyle{splncs04}
\bibliography{egbib}

\begin{thebibliography}{10}
\providecommand{\url}[1]{\texttt{#1}}
\providecommand{\urlprefix}{URL }
\providecommand{\doi}[1]{https://doi.org/#1}

\bibitem{ba2016layer}
Ba, J.L., Kiros, J.R., Hinton, G.E.: Layer normalization. arXiv preprint
  arXiv:1607.06450  (2016)

\bibitem{bijelic2020seeing}
Bijelic, M., Gruber, T., Mannan, F., Kraus, F., Ritter, W., Dietmayer, K.,
  Heide, F.: Seeing through fog without seeing fog: Deep multimodal sensor
  fusion in unseen adverse weather. In: CVPR (2020)

\bibitem{bijelic2018benchmark}
Bijelic, M., Gruber, T., Ritter, W.: A benchmark for lidar sensors in fog: Is
  detection breaking down? In: 2018 IEEE Intelligent Vehicles Symposium (IV)
  (2018)

\bibitem{brazil2019m3d}
Brazil, G., Liu, X.: M3d-rpn: Monocular 3d region proposal network for object
  detection. In: ICCV (2019)

\bibitem{caesar2020nuscenes}
Caesar, H., Bankiti, V., Lang, A.H., Vora, S., Liong, V.E., Xu, Q., Krishnan,
  A., Pan, Y., Baldan, G., Beijbom, O.: nuscenes: A multimodal dataset for
  autonomous driving. In: CVPR (2020)

\bibitem{chen2016monocular}
Chen, X., Kundu, K., Zhang, Z., Ma, H., Fidler, S., Urtasun, R.: Monocular 3d
  object detection for autonomous driving. In: CVPR (2016)

\bibitem{chen2017multi}
Chen, X., Ma, H., Wan, J., Li, B., Xia, T.: Multi-view 3d object detection
  network for autonomous driving. In: CVPR (2017)

\bibitem{chen2020monopair}
Chen, Y., Tai, L., Sun, K., Li, M.: Monopair: Monocular 3d object detection
  using pairwise spatial relationships. In: CVPR (2020)

\bibitem{ding2020learning}
Ding, M., Huo, Y., Yi, H., Wang, Z., Shi, J., Lu, Z., Luo, P.: Learning
  depth-guided convolutions for monocular 3d object detection. In: CVPR
  workshops (2020)

\bibitem{graham2017submanifold}
Graham, B., van~der Maaten, L.: Submanifold sparse convolutional networks.
  arXiv preprint arXiv:1706.01307  (2017)

\bibitem{huang2020epnet}
Huang, T., Liu, Z., Chen, X., Bai, X.: Epnet: Enhancing point features with
  image semantics for 3d object detection. In: ECCV (2020)

\bibitem{ioffe2015batch}
Ioffe, S., Szegedy, C.: Batch normalization: Accelerating deep network training
  by reducing internal covariate shift. In: ICML (2015)

\bibitem{kim2020grif}
Kim, Y., Choi, J.W., Kum, D.: Grif net: Gated region of interest fusion network
  for robust 3d object detection from radar point cloud and monocular image.
  In: IROS (2020)

\bibitem{kingma2014adam}
Kingma, D.P., Ba, J.: Adam: A method for stochastic optimization. arXiv
  preprint arXiv:1412.6980  (2014)

\bibitem{lang2019pointpillars}
Lang, A.H., Vora, S., Caesar, H., Zhou, L., Yang, J., Beijbom, O.:
  Pointpillars: Fast encoders for object detection from point clouds. In: CVPR
  (2019)

\bibitem{li2020rtm3d}
Li, P., Zhao, H., Liu, P., Cao, F.: Rtm3d: Real-time monocular 3d detection
  from object keypoints for autonomous driving. In: ECCV (2020)

\bibitem{li2022deepfusion}
Li, Y., Yu, A.W., Meng, T., Caine, B., Ngiam, J., Peng, D., Shen, J., Lu, Y.,
  Zhou, D., Le, Q.V., et~al.: Deepfusion: Lidar-camera deep fusion for
  multi-modal 3d object detection. In: Proceedings of the IEEE/CVF Conference
  on Computer Vision and Pattern Recognition. pp. 17182--17191 (2022)

\bibitem{li2022bevformer}
Li, Z., Wang, W., Li, H., Xie, E., Sima, C., Lu, T., Yu, Q., Dai, J.:
  Bevformer: Learning bird's-eye-view representation from multi-camera images
  via spatiotemporal transformers. In: ECCV (2022)

\bibitem{liang2018deep}
Liang, M., Yang, B., Wang, S., Urtasun, R.: Deep continuous fusion for
  multi-sensor 3d object detection. In: ECCV (2018)

\bibitem{lim2019radar}
Lim, T.Y., Ansari, A., Major, B., Fontijne, D., Hamilton, M., Gowaikar, R.,
  Subramanian, S.: Radar and camera early fusion for vehicle detection in
  advanced driver assistance systems. In: Machine Learning for Autonomous
  Driving Workshop at the 33rd Conference on Neural Information Processing
  Systems (2019)

\bibitem{lin2017feature}
Lin, T.Y., Doll{\'a}r, P., Girshick, R., He, K., Hariharan, B., Belongie, S.:
  Feature pyramid networks for object detection. In: CVPR (2017)

\bibitem{lin2017focal}
Lin, T.Y., Goyal, P., Girshick, R., He, K., Doll{\'a}r, P.: Focal loss for
  dense object detection. In: ICCV (2017)

\bibitem{liu2020reinforced}
Liu, L., Wu, C., Lu, J., Xie, L., Zhou, J., Tian, Q.: Reinforced axial
  refinement network for monocular 3d object detection. In: ECCV (2020)

\bibitem{ma2020rethinking}
Ma, X., Liu, S., Xia, Z., Zhang, H., Zeng, X., Ouyang, W.: Rethinking
  pseudo-lidar representation. In: ECCV (2020)

\bibitem{major2019vehicle}
Major, B., Fontijne, D., Ansari, A., Teja~Sukhavasi, R., Gowaikar, R.,
  Hamilton, M., Lee, S., Grzechnik, S., Subramanian, S.: Vehicle detection with
  automotive radar using deep learning on range-azimuth-doppler tensors. In:
  ICCV Workshops (2019)

\bibitem{manhardt2019roi}
Manhardt, F., Kehl, W., Gaidon, A.: Roi-10d: Monocular lifting of 2d detection
  to 6d pose and metric shape. In: CVPR (2019)

\bibitem{mousavian20173d}
Mousavian, A., Anguelov, D., Flynn, J., Kosecka, J.: 3d bounding box estimation
  using deep learning and geometry. In: CVPR (2017)

\bibitem{nabati2021centerfusion}
Nabati, R., Qi, H.: Centerfusion: Center-based radar and camera fusion for 3d
  object detection. In: Proceedings of the IEEE/CVF Winter Conference on
  Applications of Computer Vision (2021)

\bibitem{nobis2021radar}
Nobis, F., Shafiei, E., Karle, P., Betz, J., Lienkamp, M.: Radar voxel fusion
  for 3d object detection. Applied Sciences  (2021)

\bibitem{park2021pseudo}
Park, D., Ambrus, R., Guizilini, V., Li, J., Gaidon, A.: Is pseudo-lidar needed
  for monocular 3d object detection? In: ICCV (2021)

\bibitem{piergiovanni20214d}
Piergiovanni, A., Casser, V., Ryoo, M.S., Angelova, A.: 4d-net for learned
  multi-modal alignment. In: ICCV (2021)

\bibitem{qi2018frustum}
Qi, C.R., Liu, W., Wu, C., Su, H., Guibas, L.J.: Frustum pointnets for 3d
  object detection from rgb-d data. In: CVPR (2018)

\bibitem{qi2017pointnet++}
Qi, C.R., Yi, L., Su, H., Guibas, L.J.: Pointnet++: Deep hierarchical feature
  learning on point sets in a metric space. NeurIPS  (2017)

\bibitem{reading2021categorical}
Reading, C., Harakeh, A., Chae, J., Waslander, S.L.: Categorical depth
  distribution network for monocular 3d object detection. In: CVPR (2021)

\bibitem{richards2010principles}
Richards, M.A., Scheer, J., Holm, W.A., Melvin, W.L.: Principles of modern
  radar  (2010)

\bibitem{roddick2018orthographic}
Roddick, T., Kendall, A., Cipolla, R.: Orthographic feature transform for
  monocular 3d object detection. arXiv preprint arXiv:1811.08188  (2018)

\bibitem{ronneberger2015u}
Ronneberger, O., Fischer, P., Brox, T.: U-net: Convolutional networks for
  biomedical image segmentation. In: MICCAI (2015)

\bibitem{schumann2018semantic}
Schumann, O., Hahn, M., Dickmann, J., W{\"o}hler, C.: Semantic segmentation on
  radar point clouds. In: 2018 21st International Conference on Information
  Fusion (FUSION) (2018)

\bibitem{shah2020liranet}
Shah, M., Huang, Z., Laddha, A., Langford, M., Barber, B., Zhang, S.,
  Vallespi-Gonzalez, C., Urtasun, R.: Liranet: End-to-end trajectory prediction
  using spatio-temporal radar fusion. In: CoRL (2020)

\bibitem{sheeny2021radiate}
Sheeny, M., De~Pellegrin, E., Mukherjee, S., Ahrabian, A., Wang, S., Wallace,
  A.: Radiate: A radar dataset for automotive perception in bad weather. In:
  ICRA (2021)

\bibitem{shi2019pointrcnn}
Shi, S., Wang, X., Li, H.: Pointrcnn: 3d object proposal generation and
  detection from point cloud. In: CVPR (2019)

\bibitem{shi2020distance}
Shi, X., Chen, Z., Kim, T.K.: Distance-normalized unified representation for
  monocular 3d object detection. In: ECCV (2020)

\bibitem{simonelli2019disentangling}
Simonelli, A., Bulo, S.R., Porzi, L., L{\'o}pez-Antequera, M., Kontschieder,
  P.: Disentangling monocular 3d object detection. In: ICCV (2019)

\bibitem{simonelli2020towards}
Simonelli, A., Bulo, S.R., Porzi, L., Ricci, E., Kontschieder, P.: Towards
  generalization across depth for monocular 3d object detection. In: ECCV
  (2020)

\bibitem{srivastava2014dropout}
Srivastava, N., Hinton, G., Krizhevsky, A., Sutskever, I., Salakhutdinov, R.:
  Dropout: a simple way to prevent neural networks from overfitting. The
  journal of machine learning research  (2014)

\bibitem{srivastava2019learning}
Srivastava, S., Jurie, F., Sharma, G.: Learning 2d to 3d lifting for object
  detection in 3d for autonomous vehicles. In: IROS (2019)

\bibitem{Sun2020CVPR}
Sun, P., Kretzschmar, H., Dotiwalla, X., Chouard, A., Patnaik, V., Tsui, P.,
  Guo, J., Zhou, Y., Chai, Y., Caine, B., et~al.: Scalability in perception for
  autonomous driving: Waymo {O}pen {D}ataset. In: CVPR (2020)

\bibitem{sun2021rsn}
Sun, P., Wang, W., Chai, Y., Elsayed, G., Bewley, A., Zhang, X., Sminchisescu,
  C., Anguelov, D.: Rsn: Range sparse net for efficient, accurate lidar 3d
  object detection. In: CVPR (2021)

\bibitem{vaswani2017attention}
Vaswani, A., Shazeer, N., Parmar, N., Uszkoreit, J., Jones, L., Gomez, A.N.,
  Kaiser, {\L}., Polosukhin, I.: Attention is all you need. NeurIPS  (2017)

\bibitem{vora2020pointpainting}
Vora, S., Lang, A.H., Helou, B., Beijbom, O.: Pointpainting: Sequential fusion
  for 3d object detection. In: CVPR (2020)

\bibitem{wang2021pointaugmenting}
Wang, C., Ma, C., Zhu, M., Yang, X.: Pointaugmenting: Cross-modal augmentation
  for 3d object detection. In: CVPR (2021)

\bibitem{wang2020makes}
Wang, W., Tran, D., Feiszli, M.: What makes training multi-modal classification
  networks hard? In: CVPR (2020)

\bibitem{wang2019pseudo}
Wang, Y., Chao, W.L., Garg, D., Hariharan, B., Campbell, M., Weinberger, K.Q.:
  Pseudo-lidar from visual depth estimation: Bridging the gap in 3d object
  detection for autonomous driving. In: CVPR (2019)

\bibitem{wang2019frustum}
Wang, Z., Jia, K.: Frustum convnet: Sliding frustums to aggregate local
  point-wise features for amodal 3d object detection. In: IROS (2019)

\bibitem{weng2019monocular}
Weng, X., Kitani, K.: Monocular 3d object detection with pseudo-lidar point
  cloud. In: ICCV Workshops (2019)

\bibitem{yan2018second}
Yan, Y., Mao, Y., Li, B.: Second: Sparsely embedded convolutional detection.
  Sensors  (2018)

\bibitem{yang2020radarnet}
Yang, B., Guo, R., Liang, M., Casas, S., Urtasun, R.: Radarnet: Exploiting
  radar for robust perception of dynamic objects. In: ECCV (2020)

\bibitem{you2019pseudo}
You, Y., Wang, Y., Chao, W.L., Garg, D., Pleiss, G., Hariharan, B., Campbell,
  M., Weinberger, K.Q.: Pseudo-lidar++: Accurate depth for 3d object detection
  in autonomous driving. arXiv preprint arXiv:1906.06310  (2019)

\bibitem{zhou2019iou}
Zhou, D., Fang, J., Song, X., Guan, C., Yin, J., Dai, Y., Yang, R.: Iou loss
  for 2d/3d object detection. In: 2019 International Conference on 3D Vision
  (3DV) (2019)

\bibitem{zhou2019objects}
Zhou, X., Wang, D., Kr{\"a}henb{\"u}hl, P.: Objects as points. arXiv preprint
  arXiv:1904.07850  (2019)

\bibitem{zhou2020end}
Zhou, Y., Sun, P., Zhang, Y., Anguelov, D., Gao, J., Ouyang, T., Guo, J.,
  Ngiam, J., Vasudevan, V.: End-to-end multi-view fusion for 3d object
  detection in lidar point clouds. In: CoRL (2020)

\end{thebibliography}

\clearpage
\appendix
\section{Appendix}

We propose an efficient camera-radar sensor fusion approach for robust 3D~object detection for autonomous driving.
The method uses a ray-constrained cross-attention mechanism to leverage the range measurements from radar to improve camera depth estimates, leading to improved detection performance.
More importantly, the architecture is designed in a way that training with dropout allows the method to fall back to a single modality when one of the sensors malfunctions. \\

Here, we include more details on the following aspects:
\begin{enumerate}
  \item We describe the experimental details and present more complete results on the Waymo Open Dataset in~\ref{sec:supp_open}.
  \item We study the trade-off between latency and foreground thresholds in~\ref{sec:supp_latency}.
  \item We present ablation study on hyperparameters related to the fusion design, e.g., ray-constrained cross-attention and sensor dropout in~\ref{sec:supp_hyperparam}.
  \item We document the detailed architecture of CramNet in~\ref{sec:supp_arch}.
\end{enumerate}

\subsection{Experiment on Waymo Open Dataset}
\label{sec:supp_open}

We use the same hyperparameters as on the RADIATE dataset~\cite{sheeny2021radiate} for training a camera-only CramNet on the Waymo Open Dataset~\cite{Sun2020CVPR}.
We adopt a longer training procedure, i.e., 60k warm-up steps and 120k total steps, due to the larger size of the dataset.  We align our setting with CaDDN~\cite{reading2021categorical} to train and evaluate our performance using the front camera.
However, we train our model on a lower resolution $(640, 960)$, than in CaDDN~\cite{reading2021categorical}, $(832, 1248)$.

We report the 3D AP/APH with 0.5 and 0.7 IoU threshold on the LEVEL\_1 and LEVEL\_2 difficulties in Table~\ref{tab:supp_open}.
We conclude that our camera-only model, CramNet-C, achieves competitive performance among the state-of-the-art models.

We notice that our model performs significantly better \textcolor{red}{(+$50\% \sim 300\%$)} in the longe range region (50 m - $\infty$).
This suggests the sparse operation in 3D after the 2D segmentation filtering can better handle the long range objects, even without implicitly or explicitly modeling depth uncertainty.

\begin{table}[t]
\centering
\resizebox{1.0\linewidth}{!}{%
\begin{tabular}{c | c|c c c c|c c c c}
  \toprule
   Difficulty & Method & \ 3D AP \  & \ 0 - 30m \ & \ 30 - 50m \ & \ 50m - $\infty$\  & \ 3D APH\ & \ 0 - 30m\ & \ 30 - 50m\ & \ 50m - $\infty$\ \\
  \toprule
  \multirow{3}{*}{
  \makecell{Level 1 \\ (IoU = 0.5)}} & M3D-RPN~\cite{brazil2019m3d} & 3.79 & 11.14 & 2.16 & 0.26 & 3.63 & 10.70 & 2.09 & 0.21 \\
  & CaDDN~\cite{reading2021categorical} & 17.54 & 45.00 & 9.24 & 0.64 & 17.31 & 44.46 & 9.11 & 0.62 \\
   & \cellcolor{lightgray} CramNet-C & \cellcolor{lightgray}11.81 & \cellcolor{lightgray}32.20 & \cellcolor{lightgray}7.24 & \cellcolor{lightgray}\textbf{2.00} &
  \cellcolor{lightgray}11.59 & \cellcolor{lightgray}31.75 & \cellcolor{lightgray}7.08 & \cellcolor{lightgray}\textbf{1.93} \\
  \midrule
  \multirow{3}{*}{
  \makecell{Level 2 \\ (IoU = 0.5)}} & M3D-RPN~\cite{brazil2019m3d} & 3.61 & 11.12 & 2.12 & 0.24 & 3.46 & 10.67 & 2.04 & 0.20 \\
  & CaDDN~\cite{reading2021categorical} & 16.51 & 44.87 & 8.99 & 0.58 & 16.28 & 44.33 & 8.86 & 0.55 \\
  & \cellcolor{lightgray} CramNet-C & \cellcolor{lightgray}10.64 & \cellcolor{lightgray}30.29 & \cellcolor{lightgray}6.56 & \cellcolor{lightgray}\textbf{1.76} &
  \cellcolor{lightgray}10.44 & \cellcolor{lightgray}29.86 & \cellcolor{lightgray}6.42 & \cellcolor{lightgray}\textbf{1.69} \\
  \midrule
  \multirow{3}{*}{
  \makecell{Level 1 \\ (IoU = 0.7)}} & M3D-RPN~\cite{brazil2019m3d} & 0.35 & 1.12 & 0.18 & 0.02 & 0.34 & 1.10 & 0.18 & 0.02 \\
  & CaDDN~\cite{reading2021categorical} & 5.03 & 14.54 & 1.47 & 0.10 & 4.99 & 14.43 & 1.45 & 0.10 \\
  & \cellcolor{lightgray} CramNet-C & \cellcolor{lightgray}4.14 & \cellcolor{lightgray}\textbf{15.46} & \cellcolor{lightgray}1.20 & \cellcolor{lightgray}\textbf{0.15} &
  \cellcolor{lightgray}4.10 & \cellcolor{lightgray}\textbf{15.31} & \cellcolor{lightgray}1.19 & \cellcolor{lightgray}\textbf{0.13} \\
  \midrule
  \multirow{3}{*}{
  \makecell{Level 2 \\ (IoU = 0.7)}} & M3D-RPN~\cite{brazil2019m3d} & 0.33 & 1.12 & 0.18 & 0.02 & 0.33 & 1.10 & 0.17 & 0.02 \\
  & CaDDN~\cite{reading2021categorical} & 4.49 & 14.50 & 1.42 & 0.09 & 4.45 & 14.38 & 1.41 & 0.09 \\
  & \cellcolor{lightgray} CramNet-C & \cellcolor{lightgray}3.72 & \cellcolor{lightgray}\textbf{14.53} & \cellcolor{lightgray}1.09 & \cellcolor{lightgray}\textbf{0.13} &
  \cellcolor{lightgray}3.68 & \cellcolor{lightgray}\textbf{14.38} & \cellcolor{lightgray}1.07 & \cellcolor{lightgray}\textbf{0.13} \\
  \bottomrule
\end{tabular} }
\caption{Camera-only 3D detection results on the Waymo Open Dataset~\cite{Sun2020CVPR} validation set on the vehicle class, evaluated in terms of 3D~AP/APH at 0.5 or 0.7~IoU on the LEVEL\_1 or LEVEL\_2 difficulties.  Baseline numbers are from \cite{reading2021categorical}. Our camera-only model, CramNet-C, achieves competitive performance among state-of-the-art with the best long range detection.}
\label{tab:supp_open}
\end{table}


\subsection{Ablation Study on Latency and Foreground Threshold}
\label{sec:supp_latency}

One of the important trade-off in our model hyperparameters is the foreground segmentation threshold.
This threshold controls the density of foreground points passed from the 2D to 3D stage.
Therefore, we expect the model to perform better with a lower threshold, with the trade-off of a higher latency.

\begin{figure}
    \centering
    \includegraphics[width=0.8\linewidth]{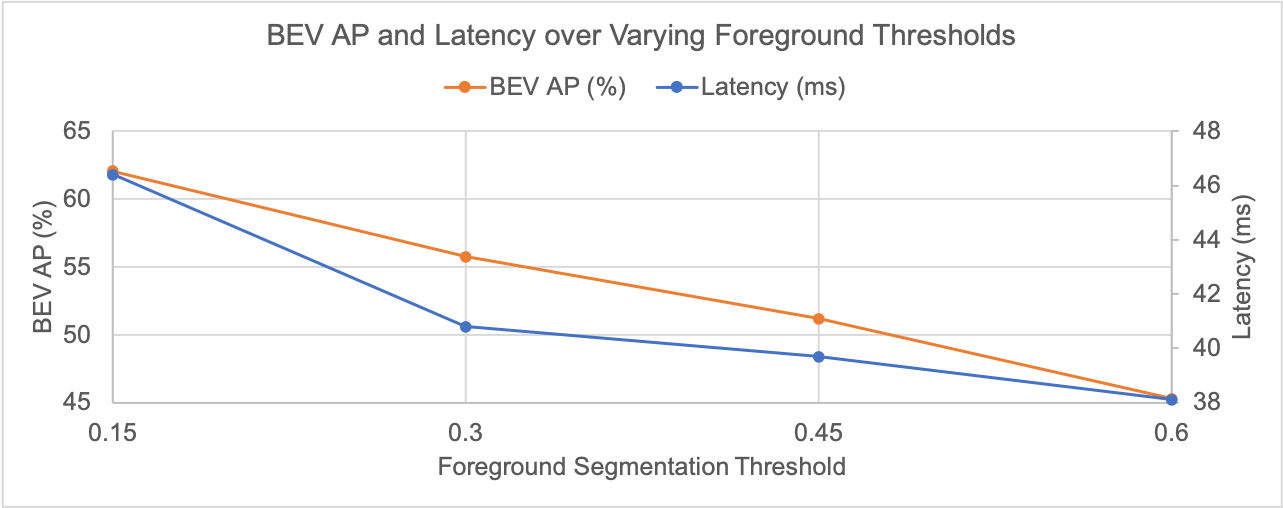}
    \caption{Ablation study on  foreground segmentation thresholds.  The model accuracy in BEV AP and latency both decreases as the foreground segmentation threshold decreases.  }
    \label{fig:supp_latency}
\end{figure}

We summarize this ablation study in Figure~\ref{fig:supp_latency}.
Our reported performance in the main paper is at the $0.15$ threshold with a latency of $46.4$ ms.
We observe a general trend of lower accuracy and lower latency when setting a lower threshold.

\subsection{Ablation Studies on Fusion Hyperparameters}
\label{sec:supp_hyperparam}

\begin{figure}
    \centering
    \includegraphics[width=\linewidth]{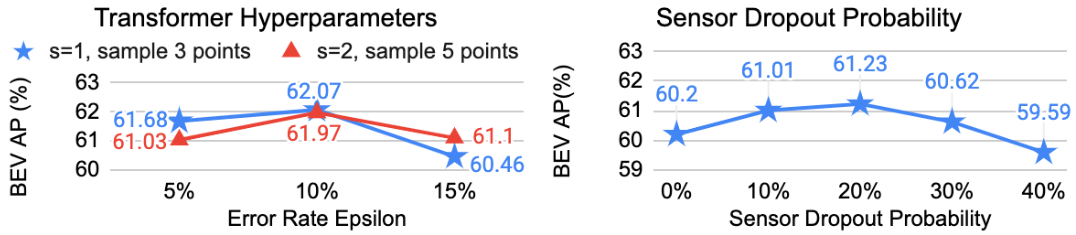}
    \caption{
    \textbf{Left}: Ablation study on hyperparameters in ray-constrained cross-attention.  
    \textbf{Right}: Abltaion study on hyperparameters in sensor dropout.  All in all, the ablation studies suggest the model is not too sensitive to hyperparameters.}
    \label{fig:supp_hyperparam}
\end{figure}

We conduct ablation studies on the effect of hyperparameters w.r.t. the 3D detection performance in BEV AP, summarized in Figure~\ref{fig:supp_hyperparam}. 
All in all, the ablation studies suggest the model is not too sensitive to hyperparameters.

For the ray constrained cross-attention, we notice the best error rate ($\epsilon=0.1$) corresponds to the general depth errors. Also, we do not need many samples along the camera ray (from 3 to 5 sampled points) as each sample already covers a large region through feature extraction.

For the sensor dropout, the performance peaks at $0.2$ dropout probability and decreases as the probability increases, indicating that too frequent dropout actually hurts the model.

For modality encoding, when removing the modality code, the BEV AP of CramNet degrades by 8.7 percentage points from $62.1\%$ to $53.4\%$.  The indicates the modality encoding is critical for the model to distinguish and utilize features from different sensors.

\subsection{Architecture Details}
\label{sec:supp_arch}

\noindent \textbf{2D U-Net.}
A downsampling block $D(B_i, C_i)$ at level $i$ contains $B$ resnet blocks with $C$-dimensional outputs, with stride $2$ in the first convolutional layer.
Each upsampling block $U(B_i, C_i)$ at level $i$ also  contains $B$ resnet
blocks with $C$-dimensional outputs. The upsampling is performed by a $1 \times 1$ convolution layer followed by a bilinear interpolation layer.
We connect the same level of the corresponding downsampling block and upsampling block to construct the U-Net.

After applying an initial $1\times 1$ convolution layer on the input with $16$-dimensional outputs, we construct a 2-D U-Net with hyperparameters specified in Table~\ref{tab:supp_2d_unet}. We use the exact same 2D U-Net for both camera and radar inputs for simplicity.  One can replace them with stronger feature extractor backbones. \\

\begin{table}[]
  \centering
  \resizebox{1.0\linewidth}{!}{%
  \begin{tabular}{|c|c|c|c||c|c|c|c|}
    \hline
    \ $D(B_1, C_1)$ \ & 
    \ $D(B_2, C_2)$ \ &
    \ $D(B_3, C_3)$ \ &
    \ $D(B_4, C_4)$ \ &
    \ $U(B_1, C_1)$ \ & 
    \ $U(B_2, C_2)$ \ &
    \ $U(B_3, C_3)$ \ &
    \ $U(B_4, C_4)$ \ \\
    \hline
    \ $(3, 16)$ \ & 
    \ $(3, 16)$ \ &
    \ $(1, 64)$ \ &
    \ $(0, 128)$ \ &
    \ $(1, 16)$ \ & 
    \ $(1, 16)$ \ &
    \ $(1, 64)$ \ &
    \ $(1, 128)$ \ \\
    \hline
  \end{tabular}}
  \caption{Detailed hyperparameters to construct a 2D U-Net.}
  \label{tab:supp_2d_unet}
\end{table}

\noindent \textbf{3D Sparse U-Net.}
We reuse the notations as 2D U-Net for 3D U-Net.
A residual block in 3D U-Net is replaced by a $3 \times 3 (\times 3)$ sparse convolution layer before the residual connection and by two $3 \times 3 (\times 3)$ submanifold sparse convolution layers within the residual block.
Unlike the symmetric downsampling and upsampling blocks in the 2D Unet, we employ 2 more downsampling blocks to output a lower resolution objectness heatmap.
We summarize the hyperparameters used to construct a 3D sparse U-Net in Table~\ref{tab:supp_3d_unet}.

\begin{table}[]
  \centering
  \resizebox{1.0\linewidth}{!}{%
  \begin{tabular}{|c|c|c|c|c||c|c|c|}
    \hline
    \ $D(B_1, C_1)$ \ & 
    \ $D(B_2, C_2)$ \ &
    \ $D(B_3, C_3)$ \ &
    \ $D(B_4, C_4)$ \ &
    \ $D(B_5, C_5)$ \ &
    \ $U(B_1, C_1)$ \ & 
    \ $U(B_2, C_2)$ \ &
    \ $U(B_3, C_3)$ \ \\
    \hline
    \ $(1, 96)$ \ & 
    \ $(2, 96)$ \ &
    \ $(2, 96)$ \ &
    \ $(1, 96)$ \ &
    \ $(1, 96)$ \ & 
    \ $(0, 96)$ \ &
    \ $(2, 96)$ \ &
    \ $(2, 96)$ \ \\
    \hline
  \end{tabular}}
  \caption{Detailed hyperparameters to construct a 3D Sparse U-Net.}
  \label{tab:supp_3d_unet}
\end{table}

\end{document}